\documentclass[subscriptcorrection,upint,varvw,mathalfa=cal=euler,balance,hyphenate,french,nocopyright]{asmejour} %

\hypersetup{%
	pdfauthor={John H. Lienhard},                       		   	
	pdftitle={ASME Journal Paper LaTeX Template},                  	
	pdfkeywords={ASME journal paper, LaTeX template, BibTeX style, asmejour class},
	pdfsubject = {Describes the asmejour LaTeX template},			
	pdflicenseurl={https://ctan.org/pkg/asmejour},
}

\JourName{Engineering for Gas Turbines and Power}

\newcommand{\p}{\Delta p_{t,\textrm{rel}}}
\newcommand{\R}{\mathbb{R}}
\newcommand{\Lo}{\mathcal{L}}

\PreprintString{GTP-24-1269}

\usepackage{multirow}

\usepackage{nomencl}

\usepackage{xstring}
\usepackage{xstring}
\usepackage{xpatch}
\patchcmd{\thenomenclature}
  {\leftmargin\labelwidth}
  {\leftmargin\labelwidth\itemindent 1em }
  {}{}
 
\newcommand{\nomenclheader}[1]{%
  \item[\hspace*{-\itemindent}\normalfont\bfseries#1]}
\renewcommand\nomgroup[1]{%
  \IfStrEqCase{#1}{%
   {R}{\nomenclheader{Roman Letters}}
   {S}{\nomenclheader{Superscripts and Subscripts}}
   {A}{\nomenclheader{Acronyms and Abbreviations}}
   {G}{\nomenclheader{Greek Letters}}
   {M}{\nomenclheader{Frequently Used Mathematical Symbols}}
   {D}{\nomenclheader{Dimensionless Groups}}
  }%
}

\makenomenclature

\usepackage{tcolorbox}
\usepackage{siunitx}
\begin{document}

\SetAuthorBlock{Patrick Krüger\CorrespondingAuthor}{Department of Mathematics,\\
   Technische Universität Berlin,\\
   Straße des 17. Juni 135,\\
   Berlin, Berlin, Germany \\
   email: krueger@math.tu-berlin.de}

\SetAuthorBlock{Hanno Gottschalk}{Department of Mathematics,\\
   Technische Universität Berlin,\\
   Straße des 17. Juni 135,\\
   Berlin, Berlin, Germany \\
   email: gottschalk@math.tu-berlin.de}

\SetAuthorBlock{Bastian Werdelmann}{Sienmens Energy,\\
   Rheinstraße 100,\\
   Mühlheim an der Ruhr, NRW, Germany \\
   email: bastian.werdelmann@siemens-energy.com} 

\SetAuthorBlock{Werner Krebs}{Sienmens Energy,\\
   Rheinstraße 100,\\
   Mühlheim an der Ruhr, NRW, Germany \\
   email: wernerkrebs@siemens-energy.com}


\title{Generative Design of a Gas Turbine Combustor Using Invertible Neural Networks}
\keywords{Combustors, Gas Turbines, Neural Networks}

\begin{abstract}
The need to burn 100\% H${}_2$ in high efficient gas turbines featuring low NO${}_x$ combustion in premix mode require the complete redesign of the combustion system to ensure stable operation without any flashback. Since all engine frames featuring a power range from \qty{4}{MW} up to \qty{ 600}{MW} are affected, a huge design effort is expected. To reduce this effort, especially to transfer knowledge between the different engine classes, generative design methods using latest AI technology will provide promising potential.
\noindent
In this work, this challenge is approached utilizing the current advances in generative artificial intelligence. We train an Invertible Neural Network (INN) on an expandable database of geometrically parameterized combustor designs with simulated performance labels. Utilizing the INN in its inverse direction, multiple design proposals are generated which fulfill specified performance labels. 
\end{abstract}

\date{Version \versionno, \today}

\maketitle 


\section{Introduction}
\begin{tcolorbox}[colback=gray!5,colframe=gray!50,boxrule=0.4pt]
\small
\textbf{Published version:}
This manuscript corresponds to the article published in
\emph{ASME Journal of Engineering for Gas Turbines and Power}, Vol.~147, Issue 1, 2025
DOI: \href{https://doi.org/10.1115/1.4066294}{10.1115/1.4066294}.
\end{tcolorbox}

In order to meet the Paris goal to limit the temperature rise below \qty{2}{K} while approaching zero CO$_2$ emission, the power production has to undergo a significant transformation. In this context, Siemens Energy takes tremendous efforts in developing highly efficient gas turbines such as the 9000HL \cite{krebs2022advanced} and new combustion systems that are capable of burning 100\% H$_2$ \cite{witzel2022development}. Both efforts are a prerequisite for a stable and economic power generation while increasing the share of regenerative power which is known for its highly fluctuating production rate. Since the reactivity of the H${}_2$/air mixture will highly increase compared to the reactivity of CH${}_4$/air mixture, a complete redesign of combustion systems of gas turbines of all sizes will be necessary \cite{witzel2022development}. 

This challenge is not easily met by conventional design procedures, where gas turbine components are designed and optimized by variation of design parameters and evaluated by numerical simulation. 'Forward' simulation assigns several performance labels to a given set of input parameters. This time-honored approach, however, requires re-iteration of simulation to conduct new parameter searches whenever the requirements have changed. Often, the process is stopped when a single design meeting the requirements has been found. Rarely, any information is given about alternative designs. Manufacturing requirements usually being not considered at this stage of the process might later trigger a re-design. All this prolongs time to market in a highly dynamic period of transition.

Negligence of data gathered in previous design efforts is another downside of traditional design methods. This changed with the arrival of surrogate models, which are widely used in gas turbine design since 2000. Based on machine learning, these models speed up the evaluation of configurations using data from previously conducted 'off-line' simulations and thereby enable better design space exploration \cite{forrester2008engineering,santner2003design}. Nevertheless, such supervised learning algorithms maintain the old logic of mapping design parameters to performance labels and therefore share many features of the traditional design approach. Further speed up can be obtained by effective, low dimensional models that serve fast data production, e.g. in conjunction with multi-fidelity optimization \cite{forrester2008engineering,zhou2022multi}.

Generative learning emerged as a new paradigm of machine learning around the middle of the last decade. While the public perceives generative learning through spectacular results in image and text generation, it has many features that are of interest in the context of design. While supervised learning usually processes a high dimensional input and predicts a few highly significant performance labels, promptable generative learning goes the other way round. On the basis of a rather low dimensional prompt, it generates multiple high dimensional outputs associated to the given labels. 

It is therefore legitimate to consider this new class of machine learning algorithms as the inverse to 'forward' supervised learning or 'forward' design. Admittedly, generative learning requires considerable amounts of data, but this, in company with foresight and planned design management, can be produced in 'off-line' simulations.
A continuous increase of the size of the simulation dataset allows training of generative models with increasing precision that suggest design alternatives from requirements with almost no latency when used 'on-line'. This sharply distinguishes generative methods from classical, simulation oriented inverse design \cite{groetsch1993inverse}. Also, as multiple designs are created from one set of requirements, manufacturing and cost can be considered in the final choice between feasible alternatives. 

A multitude of machine learning models is compatible with the above vision of generative design. Diffusion models emerged in 2015 \cite{sohl2015deep,croitoru2023diffusion} and recently revolutionized image generation. However, the success of diffusion models is based on a very large dataset of pre-trained image encodings \cite{rombach2022high}. Conditional generative adversarial networks are successfully introduced for image generation \cite{goodfellow2020generative} and for the simulation of turbulent flow \cite{kim2021unsupervised,drygala2022generative}. However, they are prone to mode collapse and poor coverage of all alternatives compatible with a given prompt \cite{gui2021review}. Normalizing flows, however, are known to be trainable on medium sized datasets and come with a good coverage. In this family belong neural ODE \cite{chen2018neural}, affine coupling flows \cite{ardizzone2018analyzing} and LU-Nets \cite{chan2023lu}. In particular, affine coupling flows have been suggested for usage of inverse design problems in \cite{ardizzone2018analyzing}.

\begin{figure}[h]
    \centering  
    \includegraphics[scale=0.47]{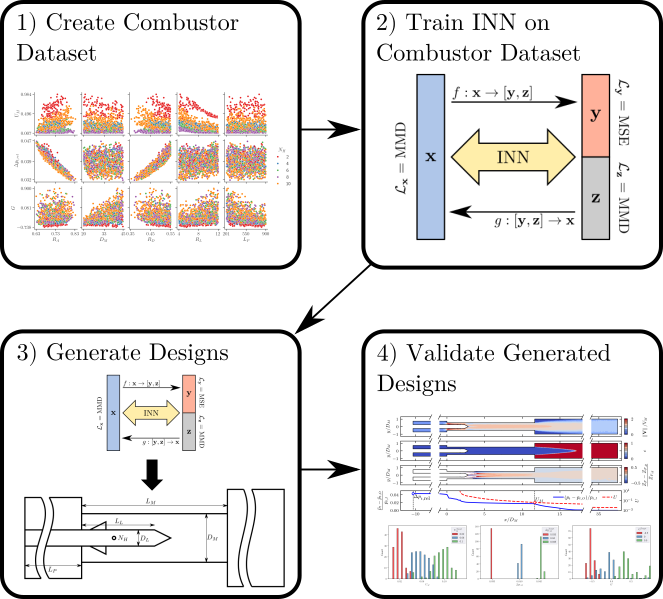}
    \caption{Generative design workflow.}
    \label{f_trainingWorkflow}
\end{figure}

In work presented in this publication, we apply the above outlined methodology to a generic configuration featuring typical gas turbine combustion items using affine coupling flow networks, specifically Invertible Neural Networks (INNs) as introduced in \cite{ardizzone2018analyzing}. In particular, this study covers the following workflow, which is illustrated in Fig. \ref{f_trainingWorkflow}:

  \begin{itemize}
     \item[1)] Create a training dataset featuring six independent geometrical parameters and three performance labels. The generation of the training dataset will be covered in Section \ref{sec_Dataset_Generation}.

     \item[2)] Train an Invertible Neural Network (INN) on basis of the training dataset which may be augmented applying surrogate models. Theoretical foundations of INNs will be covered in Section \ref{sec3}, while the practical discussion of training, augmentation and model hyperparameters will be provided in Section \ref{sec4}.

    \item[3)] Generate new designs for selected performance labels (Sections \ref{ssec_def_targets} and \ref{ssec_generate_designs}).

    \item[4)] Validate the generated designs to check the agreement with the prespecified performance labels in Section \ref{ssec_validation}. This includes a pre-validation via surrogate models and is concluded by cross-validating a selection of generated designs via Computational Fluid Dynamics (CFD) and acoustic network simulation.  

    \item[5)] Comparison with a baseline based on Gaussian processes in Section \ref{sec:comparison}.
 \end{itemize}
 
In Section \ref{sec6}, a conclusion and an outlook for future work are given.






\section{Dataset Generation}\label{sec_Dataset_Generation}

This section describes the first part of the generative training workflow, which is the creation of the dataset used to train the generative model. Each point $[\mathbf{x},\mathbf{y}]$ of the dataset is a realization of a parameterized combustor geometry defined by independent \textit{parameters} $\mathbf{x}$, which will be discussed in Section \ref{ssec_geometry}. Each realization of the parameterization is then simulated via CFD and the reacting flow network tool GeneAC (General Network Application Code) \cite{krebs2013comparison}
to obtain its corresponding \textit{performance labels} $\mathbf{y}$, which are introduced and motivated in Section \ref{sse_performance_labels}. The details of the simulation models and the boundary conditions applied, as well as the definitions of the labels are discussed in Sections \ref{ssec_cfd} and \ref{ssec_geneac} for CFD and GeneAC, respectively. Finally, an overview over the resulting dataset is given in Section \ref{ssec_dataset}.
\subsection{Geometry and Parameters}\label{ssec_geometry}

A data point in the dataset represents a realization of a parameterized combustor geometry. Compared to an actual combustor development, a strongly reduced number of independent design parameters is selected to enable a manageable amount of design points as well as the interpretation of results based on engineering judgement. The selection consists of parameters having a strong impact on the target labels and is based on combustion design experience. Furthermore, while relationships between independent parameters and target labels are still being tangible, they are ensured to be complex enough to demonstrate a satisfactory proof of concept for the generative design approach.

\begin{figure*}
    \centering  
    \includegraphics[scale=0.6]{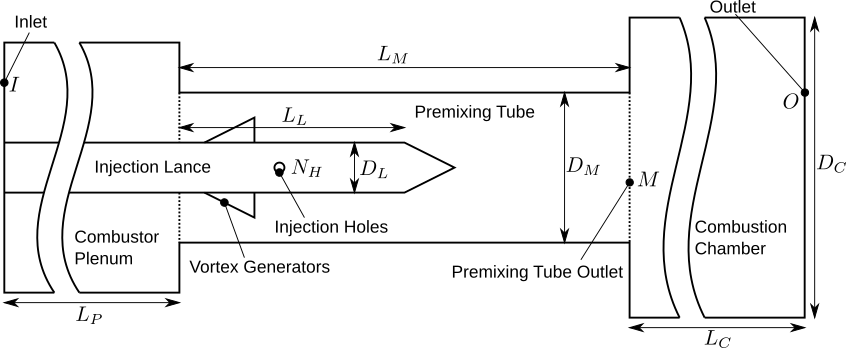}
    \caption{Overview of the parameterized combustor geometry defined by independent parameter vectors $\mathbf{x}$.}
    \label{f_geom_overview}
\end{figure*}

Each geometry consists of a combustor plenum, a premixing tube and a combustion chamber. Fuel is transported through a central fuel lance, which extends through the plenum and ends inside the premixing tube. Within the premixing tube, the lance holds several fuel holes as well as several vortex generators, which are introduced to enhance mixing. Figure \ref{f_geom_overview} gives an overview of the geometrical setup.

\subsubsection*{Independent Parameters}\label{sssec_ind_params}
Each geometry itself is uniquely defined by a vector of six independent geometrical parameters $\mathbf{x}=(R_A,N_H,D_M,R_D,R_L,L_P)$, where 
$D_M$, $R_L$ and $R_A$ are the premixing tube's diameter, length-to-diameter ratio and ratio of free area at the vortex generators to cross section area of the premixing tube, respectively. 
For the injection lance, $R_D$ is the ratio between its diameter $D_L$ and $D_M$. $N_H$ is the number of injection holes. Finally, $L_P$ is the length of the combustor plenum. An overview over $\mathbf{x}$ is given by Table \ref{t_indep_params}.
\begin{table}[t]
\caption{Independent parameters $\mathbf{x}$ of the combustor dataset.}\label{t_indep_params}%
\centering{%
\begin{tabular}{ p{0.055\textwidth} m{0.25\textwidth}  p{0.1\textwidth} }
\toprule
Parameter & \centering Description & Range \\
         \midrule
         $R_A$ & Ratio of free area at vortex generators to cross sectional area of premixing tube&$[0.63,0.83]$ \\
         $N_H$ & Number of fuel injection holes on injection lance&$[2,10]$ \\
         $D_M$& Diameter of premixing tube in \qty{}{mm}&$[20,45]$ \\
        $R_D$ & Ratio of diameter of lance to diameter of premixing tube&$[0.35,0.55]$ \\
         $R_L$ & Ratio of length of premixing tube to diameter of premixing tube&$[4,12]$ \\
         $L_P$ & Length of combustor plenum in \qty{}{mm}&$[200,900]$\\
         \bottomrule
    \end{tabular}
    }%
\end{table}

\subsubsection*{Dependent and Fixed Parameters}\label{dep_fixed_params}
All other geometrical parameters are either fixed or derived from the independent parameters $\mathbf{x}$. The relations between independent and dependent parameters as well as the values of fixed parameters have been chosen accordingly to avoid unreasonable combustor geometries. For instance, the length of the premixing tube is defined by $L_M=M_D\cdot R_L$. The diameter and length of the injection lance are derived by $D_L=R_D\cdot M_D$ and $L_L=0.2\cdot L_M$, respectively. The combustion chamber length and dump ratio are fixed at $L_C=\qty{1000}{mm}$ and $R_C=4$, respectively. The dump ratio defines the diameter of the combustion chamber $D_C$. The number of vortex generators is set to be half of the number of injection holes, $N_H$, with a minimum of 2. Some geometrical parameters are also used to derive physical boundary conditions for the CFD simulation, which will be discussed in Section \ref{ssec_cfd}.

 \subsection{Performance Labels}\label{sse_performance_labels}
 The performance labels $\mathbf{y}$ which have been selected to characterize the performance of designs are the pressure loss $\Delta p_{t,\textrm{rel}}$, the unmixedness $U_M$ at the premixing tube outlet and the thermoacoustic growth rate $G$. The pressure loss influences the gas turbine efficiency and the turbine cooling design, while the unmixedness impacts the NO$_x$ emissions. Finally, the growth rate determines if the combustion system is thermoacoustically stable. The methods which are applied to determine the performance labels are described in the following.

\subsection{CFD Simulation}\label{ssec_cfd}
In this section, the model setup and post processing of the CFD simulation are described, which is used to generate one part of the training dataset. Additionally, the labels $\mathbf{y}_{\textrm{CFD}}$ extracted from the CFD simulations are introduced.

\subsubsection*{CFD Model}\label{sssec_cfd_params}
The CFD simulations were performed using the Siemens Simcenter STAR-CCM+ software \cite{siemens2019star}. The simulation setup consists of a steady state, turbulent and reacting Reynolds Averaged Navier Stokes (RANS) model. The modeling of the turbulent internal flow is achieved through the $k$-Omega-SST model and appropriate wall functions. The gaseous fuel/air mixture, where fuel is defined as methane (CH4), is treated as compressible using the ideal gas law. The turbulent Prandtl number $P_{r,t}$ and turbulent Schmidt number $S_{c,t}$, which are needed to describe the turbulent diffusion of gas species and heat, respectively, are selected as $P_{r,t}=S_{c,t}=0.4$ in the bulk flow and are transitioned to $P_{r,t}=S_{c,t}=0.9$ towards the walls within the near wall region. Chemical reactions are modelled using a Flamelet Generated Manifold (FGM) model. The four-dimensional table consisting of mixture fraction and reaction progress variable as well as their variances was generated using the GRI3.0 mechanism. The FGM table is used to interpolate molecular fluid properties as well as the reaction rate. Turbulence effects on the reaction rate are modeled using assumed Beta PDFs for the mixture fraction and the reaction progress variable.

\subsubsection*{Computational Mesh}\label{sssec_cfd_mesh} The dominant cell type of the computational mesh is polyhedral. In the near wall region, five rows of prism layers are introduced in order to enable a range of the dimensionless wall distance between $30\leq y+ \leq 120$, which is in agreement to the requirements of the applied wall functions. The computational mesh features approximately 20 cells across the diameter of the premixing tube outlet. The same mesh resolution is applied in the flame region. Additional refinements are introduced in the vicinity of the fuel injection. Far upstream and downstream of the premixing tube, the mesh resolution is coarsened. The applied mesh settings result in cell counts ranging from 4M to 6M cells. The variation is due to changes of the dimensions of the computational domain depending on the specific sets of design parameter values being applied (e.g. length of premixing tube). Specifically, an increase of the fuel hole count increases the cell count, since many cells need to be applied in the fuel hole region to adequately resolve the fuel/air mixing process.

\begin{figure*}[t]
    \centering  
    \includegraphics[width=0.95\textwidth]{fig3.png}
    \caption{Top three charts: Normalized magnitude of flow velocity $\mathbf{V}$, progress variable $c$ and normalized deviation between local mixture fraction $Z_F$ and global mixture fraction $Z_F,g$ at the longitudinal $x$-$y$-plane of the geometry. Locations outside of the colorbar range are colored white. Bottom chart: Normalized difference between mass flow averaged total pressure $\bar{p}_t$ and the corresponding value at the domain outlet $\bar{p}_{t,O}$ and fuel/air unmixedness $U$ along the $x$-direction of the domain.}
    \label{f_flow_field}
\end{figure*}

\subsubsection*{Boundary Conditions}
Solid walls are treated with no-slip and adiabatic conditions. Additionally, the computational domain consists of two mass flow inlets, one for air and another one for fuel, as well as a pressure outlet. At the air inlet, the flow is characterized by a temperature $T_A = \qty{773.15}{K}$ and a fuel mixture fraction $Z_F = 0$. Since the created dataset contains cases with different premixing tube diameters, the air mass flow rate $\dot{m}_A$ varies depending on the considered geometry. In this context, a constraint $V_M = \qty{100}{\meter\per\second}$ was introduced on the bulk flow velocity $V_M$ at the premixing tube outlet plane $M$. Using this constraint, the required air mass flow rate 
\begin{equation}
\dot{m}_A=V_M \cdot \rho_M \cdot A_M
\end{equation}
can be approximated from continuity, where $\rho_M$ and $A_M$ are the fluid density of the fuel/air mixture and the cross-sectional area at the premixing tube outlet, respectively. 

At the fuel inlet, the flow is characterized by a temperature $T_F = \qty{473.15}{K}$ and a fuel mixture fraction $Z_F = 1$. In order to set the fuel mass flow rate $\dot{m}_F$, another constraint $Z_{F,g} = 0.03$ on the global fuel mixture fraction $Z_{F,g}$ is introduced, which is defined by the mass flow balance of the domain:
\begin{equation}
Z_{F,g} = \frac{\dot{m}_F}{\dot{m}_A+\dot{m}_F}.
\end{equation}
Accordingly, the fuel mass flow rate $\dot{m}_F$ imposed at the fuel inlet boundary can be derived:
\begin{equation}
    \Dot{m}_F= \Dot{m}_A \cdot \frac{Z_{F,g}}{1-Z_{F,g}}.
\end{equation}
At the pressure outlet of the computational domain, a static pressure of $p =$ \qty{20}{bar} is imposed.

\subsubsection*{Evaluation of an Exemplary Simulation}\label{sssec_flow_field}
In the top three charts of Fig. \ref{f_flow_field}, for an exemplary case, several flow quantities are depicted across the longitudinal $x$-$y$-plane of the geometry. The coordinates $x$ and $y$ are normalized by the premixing tube diameter $D_M$. From top to bottom, the magnitude $\left \Vert \mathbf{V} \right \Vert$ of the flow velocity $\mathbf{V}$ normalized by the premixing tube bulk flow velocity $V_M$, the reaction progress variable $c$ and the deviation of the local mixture fraction $Z_F$ from the global mixture fraction $Z_{F,g}$ normalized by $Z_{F,g}$ are shown. The latter two quantities illustrate the reaction zone and the local mixing quality, respectively. The flow velocity is moderate at the inlet plenum and accelerates significantly at the entry of the premixing tube. Further downstream, the flow expands first at the end of the fuel lance and even stronger at the inlet to the combustion chamber (the dump section). At the latter location, as can be observed from the reaction progress variable field $c$, the jet-type flame is stabilized, which is characterized by a flame length of approximately 4-5 times the premixing tube diameter $D_M$. The fuel injection is located at approximately two times the premixing tube diameter downstream of the premixing tube inlet. Within the region of approximately 1-2 times the premixing tube diameter downstream of the fuel injection, the deviation of mixture fraction from its global value is well above 50\%. In radial direction of the premixing tube, fuel rich streaks can be observed, which clearly indicate the flow path of the fuel. The deviation of the local mixture fraction continuously decreases with increasing distance from the fuel injection location.

In the bottom chart of Fig. \ref{f_flow_field}, the evolution of the total pressure $\bar{p}_t$, mass flow averaged over the cross section at the respective $x$-location, and the fuel/air unmixedness $U$, also evaluated over the cross section at the respective $x$-location, are illustrated for the same exemplary case along the $x$-direction of the domain. The averaged total pressure $\bar{p}_t$ is represented by its deviation 
\begin{equation}
\label{eq_total_pressure_drop2}
    \frac{\bar{p}_{t}-\bar{p}_{t,O}}{\bar{p}_{t,I}}
\end{equation}
from its corresponding value $\bar{p}_{t,O}$ at the domain outlet and normalized by its corresponding value $\bar{p}_{t,I}$ at the domain inlet, which illustrates the normalized total pressure loss from the respective $x$-location towards the domain outlet. In Fig. \ref{f_flow_field}, it can be observed that the most significant part of the total pressure loss is attributed to losses due to the premixing tube inlet, the jet-in-cross-flow-type fuel injection and the expanding flow field at the dump section. 

The fuel/air unmixedness $U$ is a measure for the mixing quality and is defined as a normalized and mass flow rate weighted standard deviation 
\begin{equation}
\label{eq_unmixedness}
    U = \frac{1}{Z_{F,g}} \sqrt{\frac{\int_{S}(Z_{F}-Z_{F,g})^2 \cdot \rho \mathbf{V} \cdot \mathrm{d}\mathbf{A}}{\int_{S} \rho \mathbf{V} \cdot \mathrm{d}\mathbf{A}}}
\end{equation}
of the mixture fraction $Z_{F}$ at the surface cross-section $S$ at a specific axial location $x$. In accordance to the observations with regards to the mixture fraction field, the fuel/air unmixedness is characterized by large values in the vicinity of the fuel injection and a continuous decay with increasing distance measured from the fuel injection as can be observed in Fig. \ref{f_flow_field}.

\subsubsection*{Definition of Labels}\label{sssec_eval_setup}
One part of the labels $\mathbf{y}$ of the dataset is composed of the labels $\mathbf{y}_{\textrm{CFD}}$ derived from the CFD simulations.

The labels of interest are the normalized total pressure loss $\Delta p_{t,\textrm{rel}}$ between the air inlet and the outlet of the computational domain and the fuel/air unmixedness $U_M$ at the premixing tube outlet plane.
In accordance with Equation \ref{eq_total_pressure_drop2}, the normalized total pressure loss $\Delta p_{t,\textrm{rel}}$ is defined as follows:
\begin{equation}
\label{eq_total_pressure_drop}
    \Delta p_{t,\textrm{rel}} = \frac{\bar{p}_{t,I}-\bar{p}_{t,O}}{\bar{p}_{t,I}}.
\end{equation}
In Equation \ref{eq_total_pressure_drop}, $\bar{p}_{t,S}$ denotes the mass flow averaged total pressure across the inlet plane $(S=I)$ and outlet plane $(S=O)$ of the domain. The value of $\Delta p_{t,\textrm{rel}}$ is indicated in the bottom chart of Fig. \ref{f_flow_field} for the exemplary case as a blue circle.

The fuel/air unmixedness $U_M$ at the premixing tube outlet plane $M$ is defined according to Equation \ref{eq_unmixedness}, where the surface integration is conducted over the premixing tube outlet plane ($S=M$). The value of $U_M$ is depicted in the bottom chart of Fig. \ref{f_flow_field} for the exemplary case as a red circle. 

Finally, the labels $\mathbf{y}_{\textrm{CFD}}$ derived from the CFD simulations are obtained as follows:
\begin{equation}
    \mathbf{y}_{\mathrm{CFD}} = [U_M,\p].
\end{equation}

\subsection{Acoustic Network Simulation}\label{ssec_geneac}

\subsubsection*{Model Setup}

Thermoacoustic analysis of the generic combustion system is carried out using the reacting flow network tool GeneAC. GeneAC solves the linearized Euler equations for modelling the acoustic wave propagation \cite{kru2001prediction}. The acoustic properties of the different elements including the flame are represented by transfer matrices which relate the acoustic pressure and velocity at the downstream plane to the acoustic pressure and velocity at the upstream plane of the element. Since a partially premixed jet flame is considered, equivalence ratio fluctuations and volume flow fluctuations excite the flame \cite{krebs2002thermoacoustic}. The flame transfer matrices for both volume flow fluctuations and equivalence ratio fluctuations are calculated by application of the $n-\tau$ model. At the inlet, an open acoustic end is assumed, whereas the outlet is treated as a closed end. All transfer matrices of the individual elements are cast in a homogeneous matrix equation for which solutions are obtained for discrete complex eigenfrequencies $\Omega$ \cite{krebs2013comparison}. 

\begin{figure}[h]
    \centering
    \includegraphics[scale=0.13]{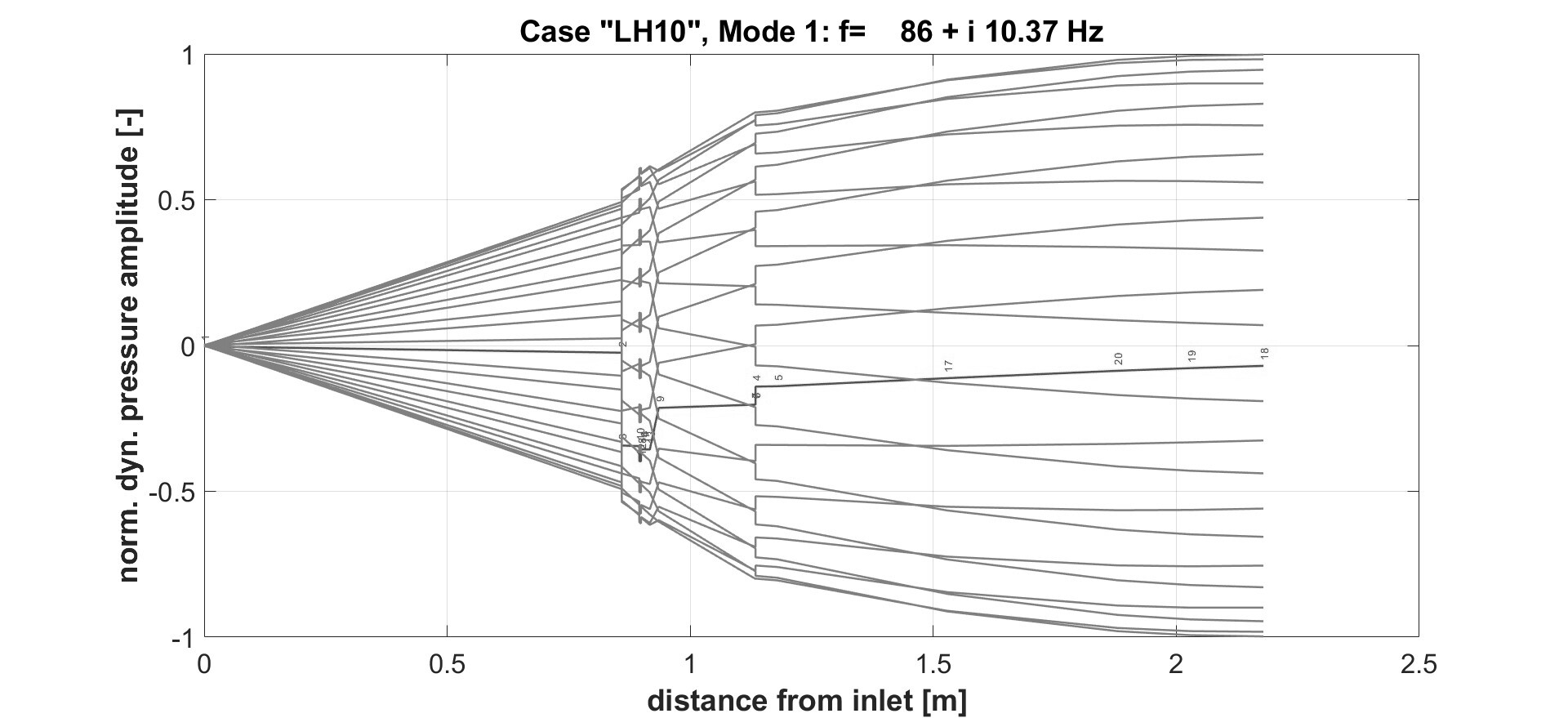}
    \caption{Quarter wave mode (f= 86~Hz) of an exemplary simulation.}
    \label{f_QWmode}
\end{figure}

\begin{figure}[h]
    \centering
    \includegraphics[scale=0.13]{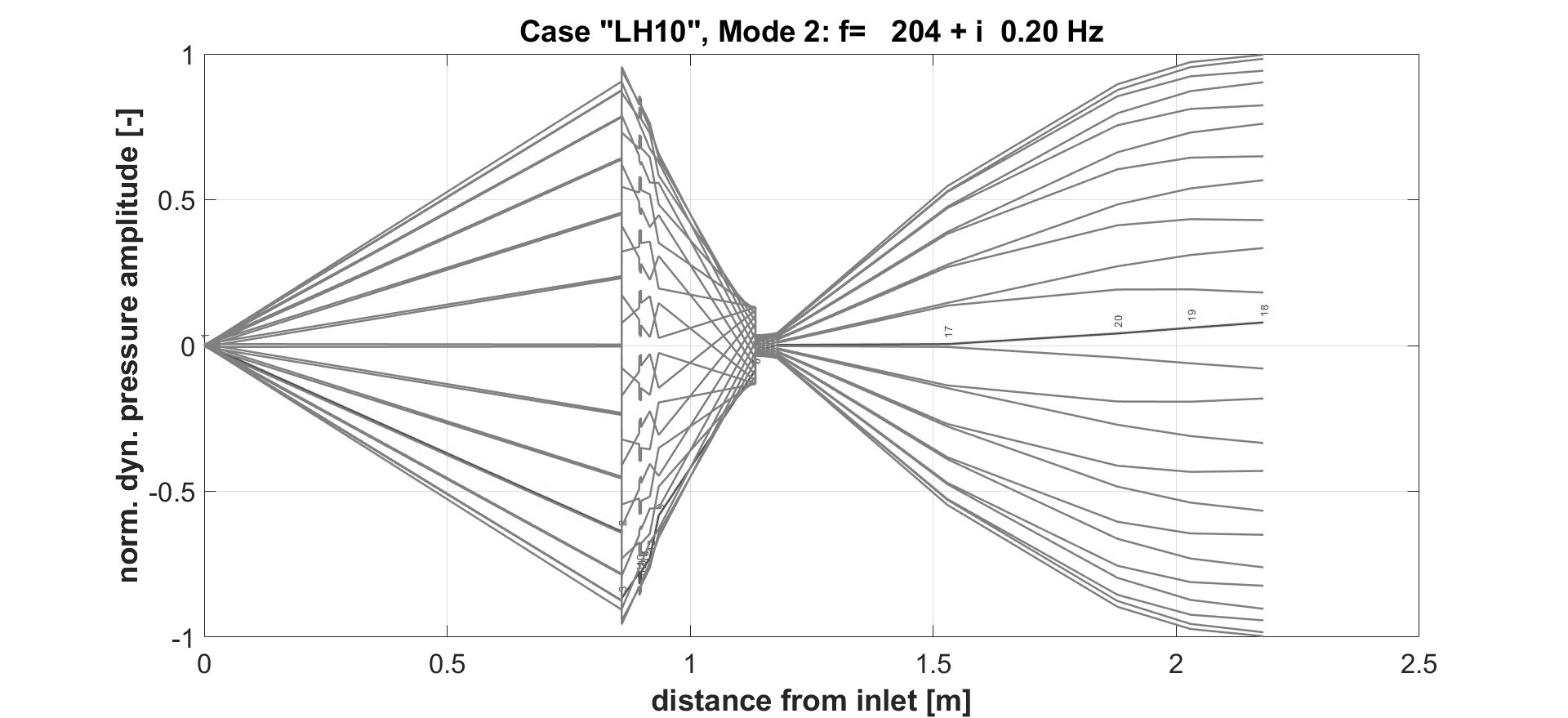}
    \caption{Three quarter wave mode (f= 204~Hz) of an exemplary simulation.}
    \label{f_3QWmode}
\end{figure}

For the system with an acoustic open inlet and a closed outlet, two acoustic modes are found in the lower frequency range: The quarter wave mode with frequencies typically below \qty{200}{Hz} and the three quarter wave mode characterized by frequencies above \qty{200}{Hz}. The actual frequency of the different modes vary from case to case due to the variation in combustor plenum length and premixing tube length. The evolution plots of the normalized dynamic pressure amplitude for both modes are shown in Fig. \ref{f_QWmode} and Fig. \ref{f_3QWmode}. For both modes, the dynamic pressure amplitude approaches zero at the open inlet and approaches one at the closed outlet. 

\subsubsection*{Definition of Labels}

The thermoacoustic system stability is represented by the thermoacoustic growth rate which results from a linear stability analysis \cite{krebs2013comparison}. The growth rate is defined as 
\begin{equation}\label{growthrate}
    G  =\frac{A_{i+1}}{A_{i}} -1 = \exp\left(-2\pi \frac{\omega_{\mathrm{imag}}}{\omega_{\mathrm{real}}}\right) -1 = \mathbf{y}_{\textrm{AC}}, 
\end{equation}    
with the complex eigenfrequency defined as 
\begin{equation}
\label{E_complexEigenfrequency}
    \Omega = \omega_{\mathrm{real}} + i * \omega_{\mathrm{imag}}.
\end{equation}    
The growth rate G expresses the ratio between the amplitude $A$ of two sequential harmonic oscillations. It is derived from the complex eigenfrequency $\Omega$ as described above. A positive value of the growth rate results in a mode excitation in which the amplitude is continuously increasing with each cycle. A negative value relates to the damping of the mode. 
For the quarter wave mode, a systematic dependence of the growth rate on the input parameters has been found, whereas for the three quarter wave mode, the dependence of the growth rate on the input parameters showed a bigger variance. Hence, since the focus was set on qualifying the generative design methodology, only the growth rate of the quarter wave is further considered. 

\begin{figure*}
    \centering
    \includegraphics[scale=0.5]{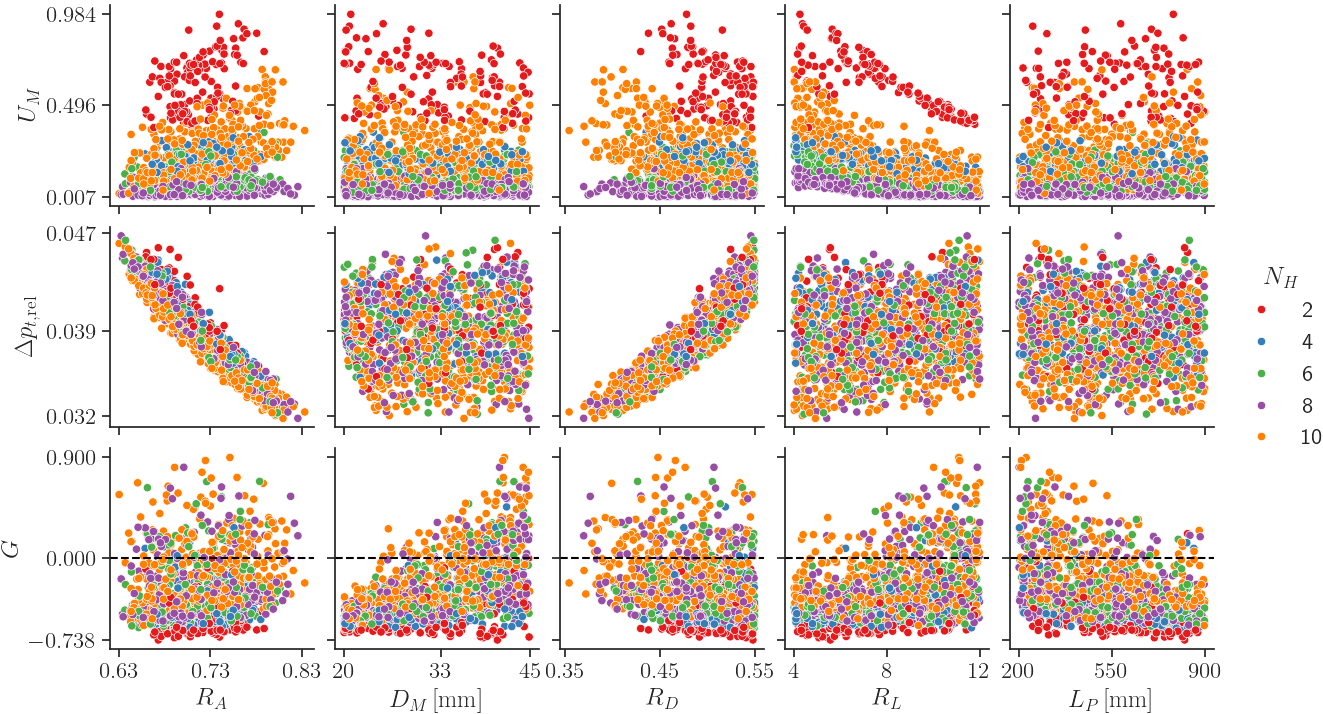}
    \caption{Scatter plot of all data points of the combustor dataset $\mathcal{D}$, displaying relationships between parameters $\mathbf{x}$ on the x-axis and labels $\mathbf{y}$ on the y-axis. The dashed horizontal line in the third row represents the boundary $G=0$ of theromacoustic stability.}
    \label{f_dataset}
\end{figure*}

\subsection{Parameter and Label Space}\label{ssec_dataset}
With the methods outlined above, the dataset $\mathcal{D}$ was created by sampling 1295 independent parameter vectors
\begin{equation}\label{eq_params}
    \mathbf{x}=[R_A,D_M,N_H,R_D,R_L,L_P].
\end{equation}
The corresponding labels $\mathbf{y}$ are composed of the labels $\mathbf{y}_{\textrm{CFD}}$ and $\mathbf{y}_{\textrm{AC}}$ obtained from the CFD and acoustic network simulations, respectively:
\begin{equation}\label{eq_labels}
    \mathbf{y} = [\mathbf{y}_{\mathrm{CFD}}, \mathbf{y}_{\mathrm{AC}}] = [U_M, \p, G].
\end{equation}
 Figure \ref{f_dataset} provides an overview plot of all sampled parameters $\mathbf{x}$ and obtained labels $\mathbf{y}$.
\subsubsection*{Design Space and Sampling}
Sampling of parameter vectors was conducted via the Latin Hypercube method \cite{mckay2000comparison} to ensure an even distribution of parameter values throughout the design space. The design space was defined by the minimum and maximum values for each independent parameter given in Table \ref{t_indep_params}.
\subsubsection*{Dataset Overview}
A wide range of values for each label results for the sampled designs, which can be seen on the y-axis of Fig. \ref{f_dataset}.
 For $\Delta p_{t,\textrm{rel}}$, strong positive and negative correlations to the ratio $R_D$ (between the diameters of the fuel lance and the premixing tube) and the ratio $R_A$ (between the free cross sectional area at the vortex generators and the premixing tube outlet) are found, respectively. A large ratio of $R_D$ represents a design with a comparably large lance diameter, which introduces a higher resistance to the flow. An inverse trend can be found for the ratio $R_A$. A high value of $R_A$ represents a design with the lowest blockage by the vortex generators and, therefore, the associated pressure loss results are on the lower side. The unmixedness $U_M$ is largely impacted by the amount of fuel holes $N_H$, where, except for $N_H=10$, the unmixedness decreases with an increase in the amount of fuel holes, which significantly improves the circumferential distribution of the fuel within the premixing tube. A minimum of $U_M$ is reached for $N_H=8$, which reveals a nonlinear relationship between $N_H$ and $U_M$. Additionally, the unmixedness decreases with increasing the length of the premixing tube, which provides obviously additional premixing length for the fuel/air mixture. The correlations between $U_M$ and the parameters $R_D$ and $R_A$ controlling the blockage of the flow by the fuel lance and the vortex generators, respectively, are inverted compared to the corresponding pressure loss correlations. This can be explained by increased velocities caused by an increase of the flow blockage, which enhances the shear layers and, hence, the production of turbulent kinetic energy. Therefore, a trade-off exists between pressure loss and unmixedness being controlled by the mentioned blockage parameters. Considering the growth rate $G$, no clear dependence was found on a single independent parameter. Instead, it was found that mode excitation depends on a more complex combination of the parameters. For example, the thermoacoustic growth rate depends on the convective time lag from fuel injection location to the flame front. Since the velocity in the premixing tube is kept constant, this value depends on the product of $D_M$ and $R_L$. Increasing the plenum length will reduce the eigenfrequency of the quarter wave mode. Although the number of unstable configurations reduces with increasing $L_P$, a stability boundary which only depends on the plenum length could not be found.

\section{Invertible Neural Neworks: Motivation and Theoretical Background}\label{sec3}

Generative design methods 'generate' new design parameter vectors $\mathbf{x}$ (Eq. \ref{eq_params}) for a specified label vector $\mathbf{y}$ (Eq. \ref{eq_labels}): 

\begin{equation}
\label{eq_generate}
    \mathscr{G} : Y \rightarrow X :  p(\mathbf{x} \in \mathbf{X} | \mathbf{y}).
\end{equation}

\begin{figure*}[h!]
    \centering
    \includegraphics[scale=0.43]{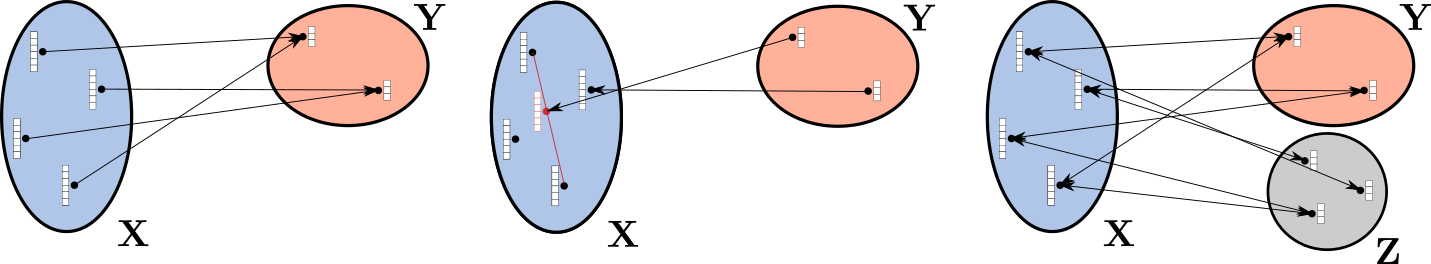}
    \caption{Different $x$ can result in the same $y$ (left), which leads to errors in direct backwards modeling (center). This is alleviated by introducing a latent variable $z$ (right).}
    \label{f_inn_motivation}
\end{figure*}

Referring to Equation \ref{eq_generate}, a generative design method $\mathscr{G}$ learns the conditional probability of design parameter vectors $\mathbf{x}\in\mathbf{X} $ given a specified target label vector $\mathbf{y}\in\mathbf{Y}$. A common problem of generative design methods is the dimensional difference between inputs and outputs. To illustrate this (Fig. \ref{f_inn_motivation}), let $\mathbf{X}\subseteq\R^P$ and $\mathbf{Y}\subseteq \R^L$ be the spaces of parameter and label vectors $\mathbf{x}$ and $\mathbf{y}$, respectively. While the forward process $\phi:\mathbf{X}\to\mathbf{Y}$ might be well understood or at least sufficiently easy to simulate, its loss of information makes it difficult to find a precise and descriptive model for the backward process $\phi^{-1}:\mathbf{Y}\to \mathbf{X}$. For instance, different parameter vectors $\mathbf{x}\in\mathbf{X}$ might yield the same label vector $\mathbf{y}\in\mathbf{Y}$ under $\phi$ (Fig. \ref{f_inn_motivation}, left). A direct modeling approach for $\phi^{-1}$ yields either only one of the possible $\mathbf{x}$, or returns an average of several $\mathbf{x}$ yielding $\mathbf{y}$, which itself might be wrong given the hidden nonlinear nature of $\phi$ (Fig. \ref{f_inn_motivation}, center). Invertible Neural Networks (INNs) approach this problem by learning a bijective mapping $f:\mathbf{X}\leftrightarrow [\mathbf{Y},\mathbf{Z}]$, where $\mathbf{Z}\subseteq\R^{P-L}$ follows a simple, easy to sample distribution $p_{\mathbf{Z}}$ 'filling up' $\mathbf{Y}$ to match the dimension of $\mathbf{X}$. $f$ then encodes information that would normally be lost in the forward process $\phi:\mathbf{X}\to \mathbf{Y}$ in $\mathbf{Z}$. This enables the simulation of $\phi^{-1}$ by running $f^{-1}$ for fixed target labels $\mathbf{y}\in \mathbf{Y}$ and varying $\mathbf{z}\in \mathbf{Z}$ to receive individual $\mathbf{x}$ yielding said $\mathbf{y}$ under $\phi$ (Fig. \ref{f_inn_motivation}, right). While this work covers a relatively low dimensional example, flow based models have been successfully applied to high dimensional datasets such as MNIST (628 dimensions) and CIFAR10 (1024 dimensions), for instance in \cite{ardizzone2019guided} and \cite{kingma2018glow}. INNs furthermore posses the universal approximation property for rather arbitrary bijective mappings. Thus, any such mapping can be approximated to arbitrary precision, see \cite{ishikawa2023universal}.

The trained INN is thus able to generate design alternatives of combustor geometries that yield specified performance values for the defined labels. The concept of INNs evolved from Normalizing Flows, which will be covered in Section \ref{ssec_normalizing_flows}. Afterwards, mathematical foundations of training as well as the special architecture of INNs will be discussed in Sections \ref{ssec_inn_training} and \ref{ssec_inn_architecture}, respectively.
\subsection{Normalizing Flows}\label{ssec_normalizing_flows}

Normalizing flows are a set of generative machine learning models that are able to learn probability distributions from given data with efficient sampling and density estimation. This is achieved by the network architecture being invertible, i.e. being able to be passed through forwards as well as backwards . This enables learning an invertible mapping between a simple distribution (e.g. a standard Gaussian) and the estimated distribution of the given data.
Let $\mathbf{Z}\subseteq \R^d$ be a set of $d$-dimensional random vectors $\mathbf{z}$ following a probability density function $p_{\mathbf{Z}}:\R^d\to \R$ that is known and easy to simulate. Let furthermore $g:\R^d\to\R^d$ be an invertible and differentiable function, let $f=g^{-1}$ be its inverse and let $\mathbf{X}=g(\mathbf{Z})$ the set of random vectors $\mathbf{x}=g(\mathbf{z})$ obtained by transforming $\mathbf{Z}$ by $g$. Then, one can use the change of variables formula (Eq. \ref{eq_change_of_vars}) to obtain the density function of $\mathbf{X}$:
\begin{equation}\label{eq_change_of_vars}
\begin{split}
    p_{\mathbf{X}}(\mathbf{x})=p_{\mathbf{Z}}(f(\mathbf{x}))\lvert  \det \mathrm{D}f(\mathbf{x})\rvert \\
    =p_{\mathbf{Z}}(f(\mathbf{x}))\lvert  \det \mathrm{D}g(f(\mathbf{x}))\rvert^{-1}.
\end{split}  
\end{equation}
Here, $\mathrm{D}f=\frac{\partial f}{\partial \mathbf{z}}$ and $\mathrm{D}g=\frac{\partial g}{\partial \mathbf{x}}$ are the Jacobians of $f$ and $g=f^{-1}$, respectively. \\
\ \\
In a generative learning setup, the function $f$ is realized as a parameterized mapping $f_{\theta}:\mathbf{X}\to \mathbf{Z}$ with learned parameters $\theta$ on a dataset $\mathcal{D}=\{\mathbf{x}^{(i)}\}_{i=1}^k$ of samples from $\mathbf{X}$. This direction of the normalizing flow is often called the \textit{normalizing direction} as the complex and unknown density $p_{\mathbf{X}}$ is transformed into the simple density $p_{\mathbf{Z}}$, which in most cases is just a multivariate standard normal distribution. To reach sufficient expressivity of the network, $f$ is set to be a composition $f=f_1\circ\cdots\circ f_n$ of individual invertible, differentiable functions $f_i,\, i\in\{1,\cdots,n\}$. The Jacobian determinant of this composition can then be calculated by 

\begin{equation}\label{eq_JacProd}
    \det \mathrm{D}f(\mathbf{x})=\prod_{i=1}^n \det \mathrm{D}f_i(\mathbf{x}_i),
\end{equation}
where $\mathbf{x_i}=f_{i+1}\circ\cdots\circ f_n(\mathbf{x})$ and thus $\mathbf{x}_n=\mathbf{x}$.
For efficient computation, the determinants of each $\mathrm{D}f_i$ need to be easy to compute. The most common way of achieving this is to design the functions $f_i$ in such a way that their Jacobians are triangular matrices, which will be further covered in Section \ref{ssec_inn_architecture}. \\
\ \\ 
The aforementioned tractabillity of the Jacobians now allows for likelihood based training and exact density estimation. Given a dataset $\mathcal{D}=\{\mathbf{x}^{(i)}\}_{i=1}^k$ sampled from $\mathbf{X}$, one can then calculate the log-likelihood of $\mathcal{D}$ under $f_{\theta}$ given the parameters $\theta$ as
\begin{equation}\label{eq_maxlikelihood}
    \begin{split}
    \log p(\mathcal{D}\rvert \theta)= \sum_{i=1}^k \log p_{\mathbf{X}}(\mathbf{x}^{(i)}\rvert\theta) \\
    = \sum_{i=1}^k \log p_{\mathbf{Z}}(f(\mathbf{x}^{(i)}\rvert \theta) + \log \lvert \det \mathrm{D}f(\mathbf{x}^{(i)}\rvert \theta)\rvert,
\end{split}
\end{equation}

with $\lvert \det \mathrm{D}f(\mathbf{x}^{(i)}\rvert \theta)\rvert$ being calculated using Eq. \ref{eq_JacProd}.
After the network has been trained using Eq. \ref{eq_maxlikelihood} or other sample-based loss functions described later, one can efficiently generate new samples following $p_{\mathbf{X}}$ of $\mathbf{X}$ by sampling from $\mathbf{Z}$ and applying $f^{-1}=g$. This is often called the \textit{generative direction} of the normalizing flow.\\
\ \\
Normalizing flows have been shown to provide precise density estimations, as well as rich and efficient sampling, for instance in \cite{dinh2016density} and \cite{kingma2018glow}. A comprehensive review of current methods is furthermore provided by \cite{kobyzev2020normalizing}.

\begin{figure}[h]
    \centering
    \includegraphics[scale=0.6]{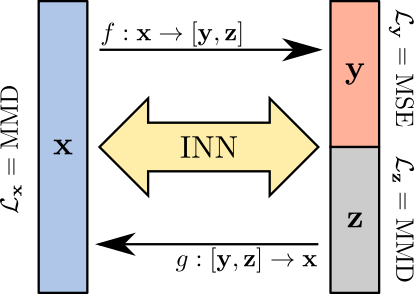}
    \caption{Training process of an INN in both directions with relevant losses.}
    \label{f_inn_concept}
\end{figure}
\subsection{INN: Training and Losses}\label{ssec_inn_training}
As illustrated in Fig. \ref{f_inn_concept}, INNs evolve the core principles of normalizing flows in two main ways: First, the data distribution $p_{\mathbf{X}}$ of $\mathbf{X}$ is not only mapped to a latent distribution $p_{\mathbf{Z}}$ of $\mathbf{Z}$. Instead, $p_{\mathbf{X}}$ is transformed into a combination of $p_{\mathbf{Z}}$ and another complex distribution $p_{\mathbf{Y}}$ of labels $\mathbf{Y}$ obtained from $\mathbf{X}$. Furthermore, INNs are trained in the normalizing direction, as well as in the generative direction in alternating fashion. 

Let $\mathcal{D}$ be a dataset consisting of parameter vectors $\mathbf{x}\in\mathbf{X}\subseteq\R^P$ and label vectors $\phi(\mathbf{x})=\mathbf{y}\in\mathbf{Y}\subseteq\R^L$ resulting from some physical process $\phi$ with $L<P$. In this work, $\mathcal{D}$ is the dataset obtained in Section \ref{sec_Dataset_Generation}, hence $P=6$ and $L=3$. Let furthermore $\mathbf{X}$ and $\mathbf{Y}$ follow complex distributions $p_{\mathbf{X}}$ and $p_{\mathbf{Y}}$. The INN is then trained in both directions. In the normalizing direction, the forward process is approximated:
\begin{equation}\label{eq_inn_forw}
    [\mathbf{y},\mathbf{z}]=f(\mathbf{x},\theta)=[f_{\mathbf{y}}(\mathbf{x},\theta),f_{\mathbf{z}}(\mathbf{x},\theta)].
\end{equation}
Here, $f_{\mathbf{y}}(\mathbf{x},\theta)$ is trained to approximate the underlying process $\phi$ by predicting labels $\mathbf{y}=\phi(\mathbf{x})$ for corresponding inputs $\mathbf{x}\in\mathcal{D}$, while $f_{\mathbf{z}}(\mathbf{x},\theta)$ is trained to follow a $(P-L)$-dimensional multivariate normal distribution $p_{\mathbf{Z}}=\mathcal{N}(0,\mathbf{1})$.

To achieve both training objectives, a combination of a supervised loss $\mathcal{L}_{\mathbf{y}}$ and an unsupervised loss $\mathcal{L}_{\mathbf{z}}$ is used for $f_{\mathbf{y}}$ and $f_{\mathbf{z}}$, respectively: 
\begin{equation}\label{eq_loss_forw}
    \mathcal{L}_{\mathrm{forw}}=\lambda_{\mathbf{y}}\mathcal{L}_{\mathbf{y}}(f_{\mathbf{y}}(\mathbf{x}^{(i)},\theta),\mathbf{y}^{(i)})+\lambda_{\mathbf{z}}\,\mathcal{L}_{\mathbf{z}}(f_{\mathbf{z}}(\mathbf{x}^{(i)},\theta),\mathbf{z}^{(i)}),
\end{equation}
where $\mathbf{z}^{(i)}$ are samples drawn from $p_{\mathbf{Z}}$.
The supervised loss function used for $\mathcal{L}_{\mathbf{y}}$ in this work is the mean squared error (MSE):
\begin{equation}\label{eq_mse}
    \mathrm{MSE}=\frac{\sum_{i=1}^k (f_{\mathbf{y}}(\mathbf{x}^{(i)},\theta)-\mathbf{y}^{(i)})^2}{k}.
\end{equation}
For $\mathcal{L}_{\mathbf{z}}$, Maximum Mean Discrepancy (MMD, see \cite{gretton2012kernel}) was applied using the inverse multiquadratic kernel $k(x,x')=\frac{1}{1+\lVert (x-x')/h\rVert}$ as in \cite{ardizzone2018analyzing}. MMD calculates the difference between the respective estimated probability distributions of two sets of input samples. Minimizing $\mathcal{L}_{\mathbf{z}}=\mathrm{MMD}(f_{\mathbf{z}}(\mathbf{x}^{(i)},\theta),\mathbf{z}^{(i)})$ thus maximizes the likelihood of $f_{\mathbf{z}}$ transforming $p_{\mathbf{X}}$ into $p_{\mathbf{Z}}$.\\

In the generative direction, the backward process $\phi^{-1}$ is trained by transforming label and noise vectors $[\mathbf{y},\mathbf{z}]$ into parameter vectors $\mathbf{x}$ yielding $\mathbf{y}$ under $\phi$:
\begin{equation}\label{eq_Inn_backaward}
    \mathbf{x}=g([\mathbf{y},\mathbf{z}],\theta).
\end{equation}

The loss function of the backward process $\Lo_{\mathrm{rev}}$ is again realized by MMD, ensuring that the distribution of generated samples $g([\mathbf{y},\mathbf{z}],\theta)$ matches $p_{\mathbf{X}}$:

\begin{equation}\label{eq_loss_backw}
\begin{split}
     \Lo_{\mathrm{rev}}=\lambda_{\mathbf{x}}\,\Lo_{\mathbf{x}}(g([\mathbf{y}^{(i)},\mathbf{z}^{(i)}],\theta),\mathbf{x}^{(i)})\\
    =\lambda_{\mathbf{x}}(\mathrm{MMD}(g([\mathbf{y}^{(i)},\mathbf{z}^{(i)}],\theta),\mathbf{x}^{(i)}).
\end{split} 
\end{equation}
In Equations \ref{eq_loss_forw} and \ref{eq_loss_backw}, scalar weights $[\lambda_{\mathbf{x}},\lambda_{\mathbf{y}},\lambda_{\mathbf{z}}]$ are introduced as hyperparameters. Depending on the application, they are adjusted to balance the different losses such that their values remain of the same order of magnitude during training. However, as only $\Lo_{\mathrm{rev}}$ and $\Lo_{\mathrm{forw}}=\lambda_{\mathbf{y}}\Lo_{\mathbf{y}}+\lambda_{\mathbf{z}}\Lo_{\mathbf{z}}$ are minimized during training, further experiments and tuning (see Section \ref{ssec_INN_Architecture_and_Tuning}) showed that a higher relative weighting of $\Lo_{\mathbf{y}}$ to $\Lo_{\mathbf{z}}$ increased predictive and thus, by invertibilty, generative accuracy as long as $\lambda_{\mathbf{z}}$ was kept just high enough to ensure that $f_{\mathbf{z}}$ (Eq. \ref{eq_inn_forw}) follows the desired normal distribution.

\subsection{INN: Invertible Architecture}\label{ssec_inn_architecture}

\begin{figure}
    \centering
    \includegraphics[scale=0.6]{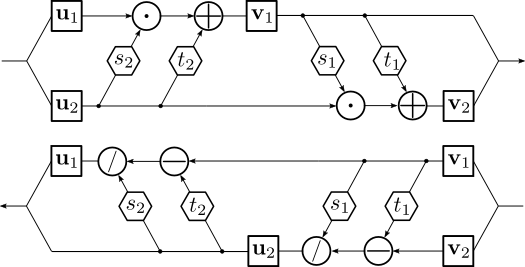}
    \caption{Illustration of operations performed by a coupling block in the forward (top) and backward (bottom) direction.}
    \label{f_coupling_blocks}
\end{figure}

To achieve invertibility, INNs consist of concatenations of so-called \textit{coupling blocks} (Fig. \ref{f_coupling_blocks}), which were first introduced by Dinh et al. \cite{dinh2016density}. Each coupling block itself consists of two affine coupling layers. The operations performed by a coupling block are as follows: First, the input $\mathbf{x}$ is split into two halves $\mathbf{u}_1=\mathbf{x}_{1:k},\mathbf{u}_2=\mathbf{x}_{k+1:n}$. The outputs of the coupling block are then computed as
\begin{equation}\label{eq_couplingblock_forw}
\begin{split}
     \mathbf{v}_1=\mathbf{u}_1\cdot \exp(s_2(\mathbf{u}_2))+t_2(\mathbf{u}_2),\\ \mathbf{v}_2=\mathbf{u}_2\cdot \exp(s_1(\mathbf{v}_1))+t_1(\mathbf{v}_1).
\end{split}
\end{equation}
Given an output $\mathbf{v}=[\mathbf{v}_1,\mathbf{v}_2]$, the input can be re-obtained as follows:
\begin{equation}\label{eq_couplingblock_backw}
\begin{split}
      \mathbf{u}_2=(\mathbf{v}_2-t_1(\mathbf{v}_1))\cdot \exp(-s_1(\mathbf{v}_1)),\\ \mathbf{u}_1=(\mathbf{v}_1-t_2(\mathbf{u}_2))\cdot \exp(-s_2(\mathbf{u}_2)).
\end{split}
\end{equation}
All operations in Equations \ref{eq_couplingblock_forw} and \ref{eq_couplingblock_backw} are performed element wise. Note here (Fig. \ref{f_coupling_blocks}) that in both directions of computation, the coefficient-defining functions $s_i,t_i,\,i\in\{1,2\}$ are passed through in the same direction, hence they themselves do not need to be invertible. Thus, these functions can be arbitrary. Furthermore, as transformations are always applied to only one half of the input at a time, the Jacobians obtained from coupling blocks are triangular, thus fulfilling the requirements for an efficient computation in e.g. Equation \ref{eq_JacProd}.

\section{INN Implementation and Tuning}\label{sec4}
In this section, the configuration of the INN as well as the tuning algorithm used to derive its most relevant hyperparameters will be discussed. Furthermore, a description and overview of surrogate models used to validate INN configurations as well as to further boost performance by data augmentation will be given. Lastly, a short overview of hardware time and resource requirements is provided. 

\subsection{INN Architecture and Tuning}\label{ssec_INN_Architecture_and_Tuning}
Important hyperparameters of the INN Architecture itself will be discussed first, followed by those relevant to the training process. Lastly, an overview of the tuning procedure will be given. 

\subsubsection*{Model Architecture}
The INN used in this work was implemented in Python using the \texttt{Framework for Easily Invertible Architectures} \cite{freia} and consists of ten coupling blocks (see Fig. \ref{f_coupling_blocks}, Eqs. \ref{eq_couplingblock_forw} and \ref{eq_couplingblock_backw}). Each of the sub-functions $s_i,t_i,\,i\in\{1,2\}$ was selected as a feed forward neural network consisting of three fully connected layers with a width of 115 neurons. Rectified Linear Units (ReLU, \cite{fukushima1975cognitron}) were applied as activation functions between layers. Between coupling blocks, invertible coordinate permutations were applied to output vectors as in \cite{kingma2018glow} and \cite{ardizzone2018analyzing} to enhance interaction of individual variables. 

\subsubsection*{Training Parameters}
The INN was trained using an Adams optimizer \cite{kingma2014adam} with a weight decay of 2e-05. Training ran over 3000 epochs with a batch size of 200 and an initial learning rate of 0.001, which was reduced by a factor of 10 after 1000 and 2000 epochs. The weights of the individual losses (see Eq. \ref{eq_loss_forw} and Eq. \ref{eq_loss_backw}) were set as $[\lambda_{\mathbf{x}},\lambda_{\mathbf{y}},\lambda_{\mathbf{z}}]=[2000,4000,400]$.

\subsubsection*{Tuning}
A hyperband algorithm \cite{li2018hyperband} was implemented using the \texttt{SMAC3} optimization package \cite{JMLR:v23:21-0888} to tune important INN and training hyperparameters which included the learning rate, batch size, number of coupling blocks, width (number of neurons) of the sub functions $s_i,t_i$ (Eqs. \ref{eq_couplingblock_forw} and \ref{eq_couplingblock_backw}), as well as the weights $[\lambda_{\mathbf{x}},\lambda_{\mathbf{y}},\lambda_{\mathbf{z}}]$ of the training losses (Eqs. \ref{eq_loss_forw} and \ref{eq_loss_backw}). Hyperband itself utilizes the random search based successive halving algorithm \cite{bergstra2012random}. This method randomly initializes hyperparameter configurations and trains them until the first of a set of user-defined budget limits, e.g. on trained epochs, is reached. Then, the worst performing configurations are cancelled in order to save computation time for better performing runs, which are continued until the next budget limit is reached. The core addition contributed by Hyperband is to start multiple succesive halving runs with different initial budget limits to avoid cancelling potentially well-performing runs that show lower performance on the first epochs too soon.\\
\ \\
To obtain a measure on the ability of different INN configurations to accurately generate designs for a wide range of targets, each trained model configuration was ran in the generative direction (see Equation \ref{eq_Inn_backaward}) to generate design parameter vectors $\{\mathbf{x}_i^{\textrm{Gen}}\}_{i=1}^n$ for a predefined set of target labels $\{\mathbf{y}_i^{\textrm{Target}}\}_{i=1}^n=\{[\p,U_M,G]_i^{\textrm{Target}}\}_{i=1}^n$:
\begin{equation}\label{eq_x_gen}
    \mathbf{x}_i^\textrm{Gen}=\mathrm{INN}^{-1}(\mathbf{y}_i^{\textrm{Target}},\mathbf{z_i}),\, \mathbf{z_i}\in\mathcal{N}(0,\mathbf{1}).
\end{equation}
 After obtaining the corresponding label values $\mathbf{y}_i^{\textrm{Gen}}$ for generated $\mathbf{x}_i^\textrm{Gen}$, the mean absolute error ($\mathrm{MAE}$) between targets and achieved values is calculated for each performance label $y\in\{U_M,\p,G\}$ individually and used as tuning objective:
 \begin{equation}\label{eq_tuning_error}
          \epsilon_y=\mathrm{MAE}(y_i^{\textrm{Target}},y_i^{\textrm{Gen}})
          =\frac{\sum_{i=1}^n |y_i^{\textrm{Target}}-y_i^{\textrm{Gen}}|}{n}.
 \end{equation}

$\mathbf{y}^{\textrm{Gen}}$ needs to be obtained for a sufficiently large number of $\mathbf{x}^\textrm{Gen}$, as well as for each of the several hundreds of tuning runs. This makes the original workflow from Section \ref{sec_Dataset_Generation} impractical due to constraints in time and computational resources. Hence, surrogate models have been introduced for fast label prediction. 

\subsection{Surrogate Models and Data Augmentation}\label{ssec_surrogate_models}

To quickly and reliably gauge the performance of the large number of INN hyperparameter configurations initialized during tuning, surrogate models were employed to emulate the forward process (Section \ref{sec_Dataset_Generation}) of deriving the performance labels $\mathbf{y}$ from design parameters $\mathbf{x}$. In the following, architecture, performance and further application of these models for data augmentation are described.
\subsubsection*{Surrogate Model Training}
A total of three surrogate models $S_{U_M},S_{\p}$ and $S_G$ were trained, each predicting the respective label from given design parameters $\mathbf{x}$. Each surrogate model was chosen as a fully connected neural network consisting of 5 layers with a width of up to 200 neurons. ReLU activation functions were again applied between layers. Training was performed on a subset $\mathcal{D}_{\textrm{train}}$ consisting of 1000 points from the original dataset $\mathcal{D}$ (see Section \ref{ssec_dataset}) using Adams optimizers. The remaining points of $\mathcal{D}$ were used as a test dataset $\mathcal{D}_{\textrm{test}}$ to validate model performance. Again, Hyperband was applied to optimize individual hyperparemeters such as learning rate and trained epochs.
\subsubsection*{Surrogate Model Performance}
The surrogate models were validated by the mean absolute errors (see Eq. \ref{eq_tuning_error}) between predicted labels 
\begin{equation}\label{eq_surrs}
    \mathbf{y}^S=[U_M^S,\p^S,G^S]=[S_{U_M}(\mathbf{x}),S_{\p}(\mathbf{x}),S_{G}(\mathbf{x})]
\end{equation}
for $n=295$ parameter vectors $\mathbf{x}\in\mathcal{D}_{\textrm{test}}$ and corresponding labels $\mathbf{y}\in\mathcal{D}_{\textrm{test}}$: 
\begin{equation}
    \epsilon_y^{S_y}=\frac{\sum_{i=1}^n |y_i^{S}-y_i|}{n},\, y\in\{U_M,\p,G\}.
\end{equation}
As shown in Table \ref{t_surrogate_error}, the average surrogate model precision is sufficient to approximate simulation results.
\begin{table}[htbp]
\caption{Mean absolute surrogate model errors for each Label.}
    \centering
    \begin{tabular}{cccc}
        \toprule
         $y$& $U_M$& $\p$& $G$\\
        \midrule
        $\epsilon_y^{S_y}$ & 0.0131 & 0.00014 &0.0251\\
      \bottomrule
    \end{tabular}
    \label{t_surrogate_error}
\end{table}
\subsubsection*{Data Augmentation}
\begin{figure}[h!]
    \centering
    \includegraphics[scale=0.53]{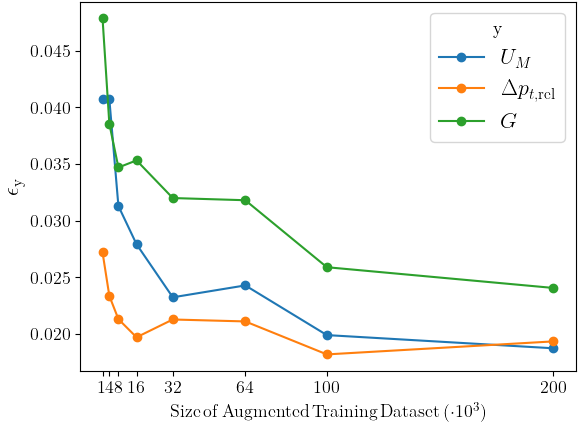}
    \caption{Errors $\epsilon_y$ (Eq. \ref{eq_tuning_error}) on individual labels by size of the augmented training dataset. The error on $\p$ was scaled by a factor of 100 for visibility.}
    \label{f_ablation}
\end{figure}

Aside from validating configurations during tuning, the surrogate models were further utilized for data augmentation, i.e. to create larger datasets $\mathcal{D}_{\textrm{aug}}$ for INN training by sampling new, larger quantities of design parameter vectors $\mathbf{x}$ and predicting the corresponding labels $\mathbf{y}=[S_{U_M}(\mathbf{x}),S_{\p}(\mathbf{x}),S_{G}(\mathbf{x})]$ instead of obtaining them through the original simulation workflow from Section \ref{sec_Dataset_Generation}. Using the tuned INN architecture obtained in Section \ref{ssec_INN_Architecture_and_Tuning}, data augmentation sizes between $2\,000$ and $200\,000$ were trained and tested for their capability of accurately generating designs for a wide range of targets. This was done as described by Equations \ref{eq_x_gen} and \ref{eq_tuning_error} with $\mathbf{y}_i^{\mathrm{Gen}}$ being predicted by the surrogate models as in Eq. \ref{eq_surrs}. The results of this ablation study are displayed in Fig. \ref{f_ablation}. Performance increases with augmentation size for the range tested, outweighing the potential noise introduced to the augmented data by the average surrogate model error (see Table \ref{t_surrogate_error}). Thus, an INN trained on an augmented dataset containing $200\,000$ points was used to conduct the validation experiments in the next section.
\begin{figure*}[h!]
    \centering
    \includegraphics[scale=0.31]{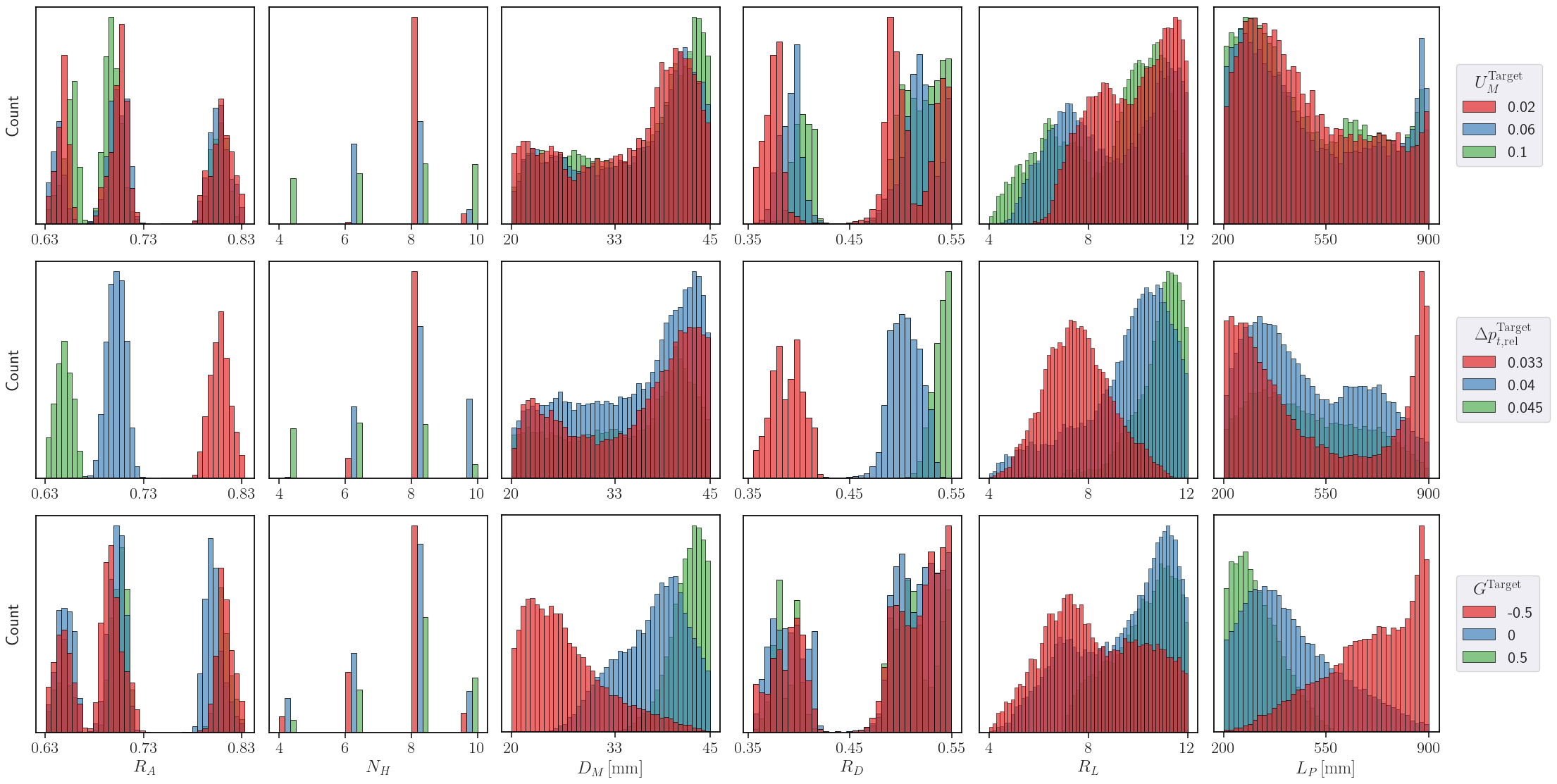}
    \caption{Distributions of the independent parameters $\mathbf{x}^{\mathrm{Gen}}$ of generated designs by target labels $\mathbf{y}^{\mathrm{Target}}$ and their respective values (top: $U_M$, center: $\p$, bottom: $G$).}
    \label{f_hist_parameters}
\end{figure*}
\subsection{Time and Resource Requirements}
INN and surrogate model training was performed on a Nvidia A100 GPU with \qty{80}{GB} of Memory. With this capacity, all three surrogate models were trained simultaneously within 20 minutes. During tuning, INNs were trained on smaller datasets containing 1000 points, again allowing for the evaluation of 10 configurations at a time within 15 minutes. Thus, while tuning can be continued indefinitely, sufficient results were obtained within 24 hours. Lastly, training the final INN configuration on the fully augmented dataset finished within 40 hours. \\
\ \\
In contrast to these efforts, CFD Simulations for the generation of the dataset (see Section \ref{sec_Dataset_Generation}) were performed on a Siemens Energy in-house HPC-Cluster with each simulation requiring an average of 96 CPU hours. Running 10 simulations in parallel using 240 CPU cores in total, the generation of all 1295 datapoints took approximately 21 days.

\section{Results}\label{sec5}
In this section, the diversity and validity of designs generated by the INN is tested by a workflow consisting of the definition of target label vectors (Sec.\ \ref{ssec_def_targets}), generation of designs for the defined targets with subsequent filtering for validity (Sec.\ \ref{ssec_generate_designs}), prevalidation using surrogate models and cross-validating a subset of prevalidated generated designs using the original dataset generation workflow (Sec.\ \ref{ssec_validation}). Finally, a comparison study using a baseline approach realized by Gaussian processes is conducted in Section \ref{sec:comparison}.

\begin{figure*}[h!]
    \centering
    \includegraphics[scale=0.28]{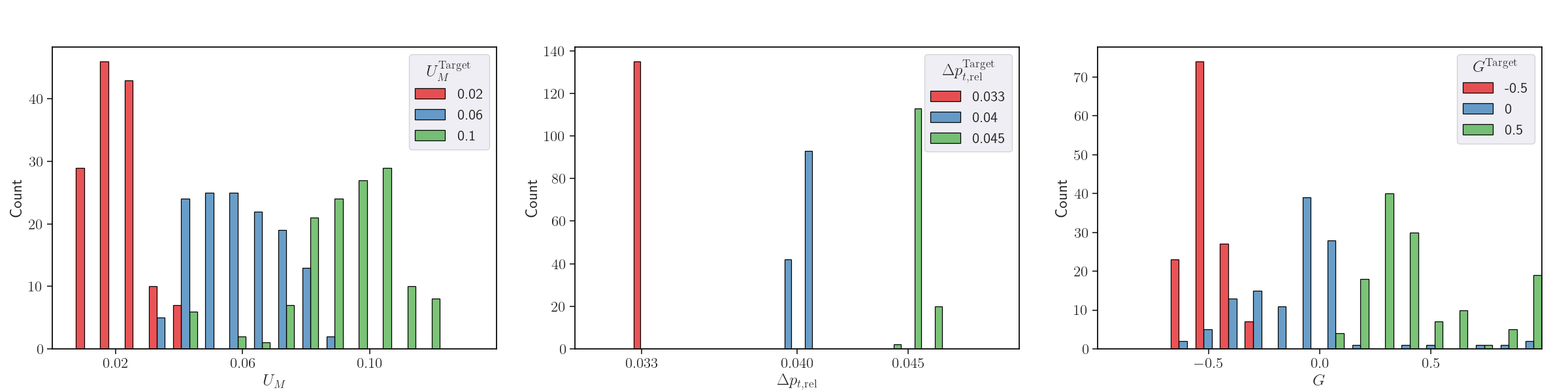}
    \caption{Distributions of labels $y^{\mathrm{Gen}}$ obtained for generated designs for target labels $y^{\mathrm{Target}}$ (left: $U_M$, center: $\p$, right: $G$), colored by target values.}
    \label{f_results}
\end{figure*}
\subsection{Definition of Targets}\label{ssec_def_targets}
A three-dimensional grid of target vectors consisting of three distinct target values for each considered performance label $[U_M,\p,G]$ was defined. For $U_M$, values of $2\%$, $6\%$ and $10\%$ were chosen. For $\p$, $3.3\%$, $4\%$ and $4.5\%$ were selected, covering the full range of this label in the underlying dataset. For $G$, target values of $-0.5$, $0$ and $0.5$ were defined to test the INN's consistency in generating both thermoacoustically stable and unstable designs as well as to investigate the behavior on the stability boundary. The chosen target label values are listed in Table \ref{t_grid_targets} and result in 27 individual target label vectors $\mathbf{y}_i^{\textrm{Target}}$.

\begin{table}[htbp]
\caption{Grid points for target label vectors used for validation.}
\centering
\begin{tabular}{cccc}
\toprule
Label & \multicolumn{3}{c}{Target Values}  \\ 
\midrule
$U_M$& 0.02 & 0.06 & 0.1 \\
$\p$ & 0.033 & 0.04 & 0.045\\ 
$G$ & -0.5 & 0 & 0.5\\
\bottomrule
\end{tabular}
\label{t_grid_targets}
\end{table}

\subsection{Generation and Filtering of Designs}\label{ssec_generate_designs}
For each $\mathbf{y}_i^{\textrm{Target}},\,i\in\{1,\cdots,27\}$, a set of $n=5000$ parameter vectors $\{\mathbf{x}_j^{\textrm{Gen},i}\}_{j=1}^{n}$ was generated using the INN. Generated designs were then filtered to remain within the boundaries of the initial dataset (Fig. \ref{f_dataset}) for each respective independent parameter, with an average yield rate of $60\%$ for valid deigns. Figure \ref{f_hist_parameters} gives an overview over the distributions of individual parameters of designs generated for each target value (represented by colors) of the individual labels $U_M$ (top), $\p$ (center) and $G$ (bottom). For instance, the green distributions in the top row correspond to the independent parameters of all generated designs $\mathbf{x}^{\mathrm{Gen},i}$ for all target vectors $\mathbf{y}_i^{\mathrm{Target}}$ that have a target unmixedness of $U_M^\mathrm{Target}=0.1$ while having any of the respective predefined values (Table \ref{t_grid_targets}) for $\p^{\mathrm{Target}}$ and $G^\mathrm{Target}$.\\

A variety of different alternatives of design parameters $\mathbf{x^{\mathrm{Gen}}}$ can be realized for a given choice of target labels. $\p$ largely determines the distributions of generated parameters $R_A$ (first column, center) and $R_D$ (fourth column, center) while having no significant influence on the distribution of $D_M$. $N_H$ is primarily influenced by the choice of $U_M$. A target value of 2\% almost exclusively generates $\mathbf{x}^{\mathrm{Gen}}$ with 8 injection holes, while a target value of $U_M= $ 10\% yields an even distribution of designs between $N_H=4$ and $N_H=8$. The choice of $G$ has an impact on generated values for the plenum length $L_P$ and the diameter of the premixing tube $D_M$ which itself is correlated to the length of the premixing tube $L_M$ via the parameter $R_L$. The observed dependencies between target labels and generated parameter distributions are in agreement with the physical relationships between parameters and labels discussed in Section \ref{ssec_dataset}.

\subsection{Validation}\label{ssec_validation}
Each individual filtered Set $\{\mathbf{x}_j^{\textrm{Gen},i}\}$ was prevalidated as in Eq. \ref{eq_tuning_error} using the surrogate models introduced in Section \ref{ssec_surrogate_models}, identifying the subset of the 20\% best performing designs. From each of the identified subsets, a selection of $k=15$ designs $\{\mathbf{x}_j^{\textrm{Gen},i}\}_{j=1}^k$ was made while maximizing the Euclidean distance between the independent parameters of subsequent choices to ensure maximal diversity. For the resulting Set $\{\mathbf{x}_j^{\textrm{Gen},i}\mid i\in\{1,\cdots,27\},\,j\in\{1,\cdots,15\}\}$ of 405 parameter vectors, the true label vectors $\mathbf{y}_{i,j}^{\textrm{Gen}}$ were obtained using the original simulation workflow from Section \ref{sec_Dataset_Generation}. The obtained label values $y_{i,j}^{\mathrm{Gen}},\,y\in\{U_M,\p,G\}$ are displayed via the histogram plots in Fig. \ref{f_results}, which are divided analogously to Fig. \ref{f_hist_parameters} by individual labels $U_M$ (left), $\p$ (center) and $G$ (right) and colored by the individual target values for each label. Table \ref{t_results} furthermore gives a numerical evaluation by providing mean $\mu(y)$ and standard deviation $\sigma(y)$ of the obtained $\mathbf{y}_{i,j}^{\textrm{Gen}}$ for each target value of each respective label.\\

\begin{table}[h!]
\caption{Mean $\mu$ and standard deviation $\sigma$ for true label values $y^{\mathrm{Gen}}$ obtained from generated designs for each target label value $y^{Target}$.}
\centering
\begin{tabular}{cccc}
\toprule
$y$ & $y^{\mathrm{Target}}$ & $\mu(y^{\mathrm{Gen}})$ & $\sigma(y^{\mathrm{Gen}})$ \\ 
\midrule
\multirow{3}{*}{$U_M$}& 0.02&0.0222&0.0078\\
 &0.06 & 0.0572& 0.0131\\
 &0.1&0.0916&0.0165\\

\multirow{3}{*}{$\p$}&0.033&0.03306&0.00014\\
&0.04&0.04011&0.00015\\
&0.045&0.0453&0.00034\\

\multirow{3}{*}{$G$}&-0.5&-0.4938&0.0826\\
&0&-0.0959&0.2618\\
&0.5&0.4415&0.2765\\
\bottomrule
\end{tabular}
\label{t_results}
\end{table}

$\p$ is the best performing label yielding high accuracy of generated designs. This was expected due to the strong correlations between parameters and $\p$ present in the dataset, but also due to the high accuracy of the pressure loss prediction using the applied CFD methods. More uncertainty is introduced with $U_M$, with most designs falling within a deviation of $\pm 2\%$ around the respective target value. The highest uncertainty is observed with $G$. Here, designs targeting a value of $-0.5$ all remain negative, albeit with higher spread of up to $0.27$ in both directions. For $G=0$, the spread increases further, with results ranging between $-0.56$ and outliers above $1$, while the majority of cases remains centered around the target label value. Further disagreement between targets and results is observed for $G=0.5$, which may be due to lower prevalence of unstable designs in the dataset. However, all evaluated designs remain of positive growth rate, hence the main objective of reliably generating either thermoacoustically stable or unstable designs can be viewed as fulfilled. In Fig. \ref{f_results} and Table \ref{t_results}, 5 \% of generated designs that yielded outlier values of $G^{\mathrm{Gen}}>1$ were clipped to a value of 1. 
The errors found between target label values and generated label values are composed both of the error imposed by the INN and the errors of the prediction methods. The significantly higher prediction errors for the unmixedness and the growth rate result in wider distributions for these labels seen in Fig. \ref{f_results}.
\subsection{Comparison to Gaussian Process Baseline}\label{sec:comparison}

Gaussian processes (GP) are a versatile machine learning tool that is widely used in gas turbine optimization \cite{forrester2008engineering}. In contrast to invertible models, GPs only learn the forward direction. The inverse problem can be approximated by
\begin{equation}\label{eq_gp}
    \min_{\mathbf{x}\in\mathbf{X}}\lVert \mathbf{y}-\mathrm{GP}(\mathbf{x})\rVert^2
\end{equation}
and solved by numerical optimization. In this baseline study, the Nelder-Mead pseudo-simplex algorithm \cite{nelder1965simplex} was used. To ensure diversity of retrieved designs, starting points $x_0$ for the optimization problem (Eq. \ref{eq_gp}) are sampled from a uniform distribution on $\mathbf{X}$. Here, $\mathrm{GP}(\mathbf{x})$ stands for three distinct GP models trained to predict each respective label $y\in\{ U_M,\p,G\}$. \\
\ \\
The GPs have been fitted using the the standard settings of the \texttt{scikit-learn} Python3 library. The Nelder-Mead minimizer was taken from the same library. In all cases, convergence was reached within one to three seconds prior to the maximum number of iterations, which was set to 1000. \\

With this method, $n=2000$ designs $\{\mathbf{x}_j^{\textrm{GP},i}\}_{j=1}^{n}$ were obtained for each $\mathbf{y}_i^{\textrm{Target}}$ defined in Section \ref{ssec_def_targets} and filtered as described in Section \ref{ssec_generate_designs}, with an average valid design yield of 50\%. In analogy to Section \ref{ssec_validation}, designs for each $\mathbf{y}_i^{\textrm{Target}}$ were prevalidated by the fitted GP models to identify the 20\% best performing designs, for which the true label vectors $\mathbf{y^{\mathrm{GP}}}$ were then obtained via the surrogate models (Section \ref{ssec_surrogate_models}).\\

Table \ref{t_gp_results} provides the mean absolute errors (see Eq. \ref{eq_tuning_error}) $\epsilon^{\mathrm{GP}}=\mathrm{MAE}(y^{\mathrm{Target}},y^{\mathrm{GP}})$ of the baseline model and $\epsilon^{\mathrm{INN}}=\mathrm{MAE}(y^{\mathrm{Target}},y^{\mathrm{Gen}})$ for the INN approach, displayed again for each target value of each label individually, as well as their relative difference $\Delta=\frac{\epsilon^{\mathrm{GP}}-\epsilon^{\mathrm{INN}}}{\epsilon^{\mathrm{GP}}}$ in percent. As is visible, the GP baseline model is outperformed by the INN approach on all target label values, with $\Delta$ ranging between $13.55\%$ for $G=0.5$ and $94.85\%$ for $U_M=0.02$. The big difference between the obtained solutions is a consequence of the non-local nature of the inverse problem, which leads the Nelder-Mead descent algorithm to be caught in local minima of the GPs. Global optimization algorithms can potentially overcome this problem; they would, however, be at risk of undergoing a mode collapse and thus not producing sufficiently diverse designs.

\begin{table}[h!]
\caption{Mean absolute errors between target and obtained label values for Gaussian process baseline ($\epsilon^{\mathrm{GP}})$ and INN ($\epsilon^{\mathrm{INN}})$, as well as the relative difference between the former and the latter. Slight change from previous publication due to correction of code error.}
\centering
\begin{tabular}{ccccc}
\toprule
$y$ & $y^{\mathrm{Target}}$ & $\epsilon^{\mathrm{GP}}$ &$\epsilon^{\mathrm{INN}}$& $\Delta$ (\%)\\ 
\midrule
\multirow{3}{*}{$U_M$}& 0.02&0.1262&0.0065&94.85\\
 &0.06 & 0.0915& 0.0113&87.65\\
 &0.1&0.0828&0.0134& 83.82\\

\multirow{3}{*}{$\p$}&0.033&0.0010&0.0001&90.00\\
&0.04&0.0013&0.0002&84.62\\
&0.045&0.0041&0.0004&90.24\\

\multirow{3}{*}{$G$}&-0.5&0.0916&0.0632&31.00\\
&0&0.6172&0.2456&60.20\\
&0.5&0.2163&0.1870&13.55\\
\bottomrule
\end{tabular}
\label{t_gp_results}
\end{table}

A further advantage of the generative workflow lies in the computation time during model inference. For the GP baseline, each sampled starting point $x_0$ and hence any generated design requires a separate solving of (Eq. \ref{eq_gp}) taking about one second. Thus, the production of a sufficient number of design alternatives for a given set of labels $\mathbf{y}^{\mathrm{Target}}$ can take up to an hour, whereas the generation of designs with $\mathrm{INN}$s is instantaneous and therefore suited for interactive design meetings if combined with fast 3D visualization. Additionally, this allows to further enhance diversity by generating designs for ranges $y^{\mathrm{Target}}_1<\cdots < y^{\mathrm{Target}}_n$ when only the fulfillment of certain inequality constraints, e.g. $G<0$, is required. $\mathrm{INN}$s furthermore guarantee the continuous change of the designs with continuous changes in the requirements, provided the noise vector is held fixed, while this is not the case for inverse problems on GPs. This favorable property of $\mathrm{INN}$s supports the understanding of design principles by engineers.

Thus, for the specific purpose of inverse design and the use case considered here, $\mathrm{INN}$s clearly outperform the GP baseline. Certainly, there are other domains of application, like surrogate based optimization, where GPs presumably are superior to $\mathrm{INN}$s.

\section{Conclusion and Outlook }\label{sec6}
This work demonstrated a generative design workflow using Invertible Neural Networks. The workflow has been applied to a generic combustion system and is able to produce diverse design alternatives which meet predefined performance requirements. The dataset on which the generative design method is trained consists of six design parameters and three associated performance labels which were produced with CFD as well as the reacting flow network tool GeneAC. The dataset was further augmented using surrogate models. The methodology can be easily expanded to include more design parameters and labels. The performance labels can also be produced by experiments instead of computations, or by a mixture of both. Generally, the design methodology aims at building up a large training data space which will be continuously expanded and will automatically include lessons learnt from past design efforts. It will accelerate the design process for generating new combustion systems which are needed to fuel 100\% H${}_2$. 
The presented generative design methodology differentiates to Multi Disciplinary Optimization (MDO) by issuing a variety of different potential design solutions rather than one optimal solution. The set of potential solutions can then be further narrowed down in consecutive design steps like manufacturability reviews. Since the generative design methodology generates new design solutions, they can be used to further extend the training dataset and hence establish an active learning approach.

As a next step, the current methodology is applied to the design of the premixing passage of burners capable of 100\% H${}_2$. Free form design elements are used to represent the complex shapes of the purge system. In this context, the flashback propensity is introduced as an additional performance parameter that will be obtained both through experiments and computations.

\section*{Acknowledgments}
The authors acknowledge financial support through grant no. 03EE5069H by the Federal Ministry of Economic Affairs and Climate Action (Germany) and by Siemens Energy.

\section*{Funding Data}

\begin{itemize}
\item German Federal Ministry of Economic Affairs and Climate Action (Grant No.\ 03EE5069H)
\end{itemize}


\printnomenclature

\nomenclature[A]{$\mathrm{FGM}$}{Flamelet Generated Manifold}
\nomenclature[A]{$\mathrm{INN}$}{Invertible Neural Network}
\nomenclature[A]{$\mathrm{MAE}$}{Mean absolute error}
\nomenclature[A]{$\mathrm{MSE}$}{Mean squared error}
\nomenclature[A]{$\mathrm{MMD}$}{Maximum mean discrepancy}
\nomenclature[A]{$\mathrm{GP}$}{Gaussian process}
\nomenclature[A]{$\mathrm{GeneAC}$}{General Network Application Code}
\nomenclature[A]{$\mathrm{PDF}$}{Probability density function}

\nomenclature[R,02]{$A$}{Cross-sectional area [\qty{}{\meter^2}]}
\nomenclature[R,03]{$A$}{Amplitude of pressure oscillation [\qty{}{\kg\m^{-1}\s^{-2}}]}
\nomenclature[R,10]{$R$}{Ratio}
\nomenclature[R,06]{$L$}{Length [\qty{}{\meter}]}
\nomenclature[R,04]{$D$}{Diameter [\qty{}{\meter}]}
\nomenclature[R,07]{$N$}{Count}
\nomenclature[R,11]{$T$}{Temperature [\qty{}{\K}]}
\nomenclature[R,17]{$Z$}{Mixture fraction}
\nomenclature[R,14]{$V$}{Bulk flow velocity [\qty{}{\meter\s^{-1}}]}
\nomenclature[R,05]{$G$}{Thermoacoustic growth rate}
\nomenclature[R,12]{$U$}{Unmixedness}
\nomenclature[R,16]{$y_+$}{Dimensionless wall distance}
\nomenclature[R,15]{$x$, $y$}{Cartesian coordinates [\qty{}{\meter}]}
\nomenclature[R,08]{$p$}{Static pressure [\qty{}{\kg\m^{-1}\s^{-2}}]}
\nomenclature[R,07]{$\Dot{m}$}{Mass flow rate [\qty{}{\kg\s^{-1}}]}
\nomenclature[R,09]{$\bar{p}_t$}{Mass flow averaged total pressure [\qty{}{\kg\m^{-1}\s^{-2}}]}
\nomenclature[R,13]{$\mathbf{V}$}{Velocity vector [\qty{}{\meter\s^{-1}}]}
\nomenclature[R,01]{$\mathbf{A}$}{Area vector [\qty{}{\meter^2}]}

\nomenclature[G]{$\rho$}{Fluid density [\qty{}{\kg \meter^{-3}}]}
\nomenclature[G]{$\Delta$}{Difference}
\nomenclature[G]{$\Omega$}{Complex eigenfrequency [\qty{}{\s^{-1}}]}
\nomenclature[G]{$\omega$}{Part of complex eigenfrequency [\qty{}{\s^{-1}}]}

\nomenclature[D]{$P_{r,t}$}{Turbulent Prandtl number}
\nomenclature[D]{$S_{c,t}$}{Turbulent Schmidt number}

\nomenclature[S,09]{$M$}{Premixing tube and premixing tube outlet}
\nomenclature[S,11]{$P$}{Combustor plenum}
\nomenclature[S,02]{$C$}{Combustion chamber}
\nomenclature[S,08]{$L$}{Fuel injection lance}
\nomenclature[S,05]{$H$}{Holes}
\nomenclature[S,03]{$F$}{Fuel}
\nomenclature[S,01]{$A$}{Air}
\nomenclature[S,14]{$S$}{Arbitrary surface}
\nomenclature[S,10]{$O$}{Domain outlet}
\nomenclature[S,06]{$I$}{Domain inlet}
\nomenclature[S,04]{$g$}{Global}
\nomenclature[S,13]{rel}{Relative}
\nomenclature[S,12]{$\mathrm{real}$}{Real part of complex number}
\nomenclature[S,07]{$\mathrm{imag}$}{Imaginary part of complex number}

\nomenclature[M]{$\mathbf{X}/\mathbf{Y}/\mathbf{Z}$}{Parameter/label/noise vector space}
\nomenclature[M]{$\mathbf{x}/\mathbf{y}/\mathbf{z}$}{Singular parameter/label/noise vector}
\nomenclature[M]{$x/y$}{Scalar value of specific parameter/label}
\nomenclature[M]{$p_{\mathbf{X}/\mathbf{Y}/\mathbf{Z}}$}{Distributions of parameter/label/noise vectors}
\nomenclature[G]{$\epsilon$}{Error}

\selectlanguage{english} 



\nocite{*} 

\bibliographystyle{asmejour}   

\bibliography{GTP-24-1269} 



\end{document}